\documentclass[10pt,twocolumn,letterpaper]{article}

\usepackage{cvpr}
\usepackage{times}
\usepackage{epsfig}
\usepackage{graphicx}
\usepackage{amsmath}
\usepackage{amssymb}
\usepackage{epigraph}
\usepackage{mathtools}
\usepackage{flushend}
\usepackage{makecell}
\usepackage{multirow}
\usepackage{booktabs}
\usepackage[table,xcdraw]{xcolor}
\DeclarePairedDelimiter{\ceil}{\lceil}{\rceil}

\usepackage[pagebackref=true,breaklinks=true,letterpaper=true,colorlinks,bookmarks=false]{hyperref}

\cvprfinalcopy 
\def\cvprPaperID{8535} 

\ifcvprfinal\fi

\newcolumntype{L}[1]{>{\raggedright\let\newline\\\arraybackslash\hspace{0pt}}m{#1}}
\newcolumntype{C}[1]{>{\centering\let\newline\\\arraybackslash\hspace{0pt}}m{#1}}
\newcolumntype{R}[1]{>{\raggedleft\let\newline\\\arraybackslash\hspace{0pt}}m{#1}}

\begin{document}

\title{Dynamic Spatial Verification for Large-Scale Object-Level Image Retrieval}

\author{Joel Brogan$^1$ \and Daniel Moreira$^1$ \and Aparna Bharati$^1$ \and Kevin Bowyer$^1$ \and Patrick Flynn$^1$ \and Walter Scheirer$^1$\\
$^1$Department of Computer Science and Engineering,\\ University of Notre Dame, US\\
{\tt\small jbrogan4,dhenriq1,abharati,}\\
{\tt\small kwb,flynn,walter.scheirer@nd.edu}
\and
Anderson Rocha$^2$\\
$^2$Institute of Computing\\
University of Campinas, Brazil\\
{\tt\small anderson.rocha@ic.unicamp.br}
}

\maketitle

\begin{abstract}
Images from social media can reflect diverse viewpoints, heated arguments, and expressions of creativity, adding new complexity to 
retrieval tasks.
Researchers working on Content-Based Image Retrieval (CBIR) have traditionally tuned their 
algorithms to match filtered results with user search intent.
However, we are now bombarded with composite images of unknown origin, authenticity, and even meaning.
With such uncertainty, users may not have an initial idea of what the results of a search query should look like.
For instance, hidden people, spliced objects, and subtly altered scenes can be difficult for a user to detect initially in a meme image, but may contribute significantly to its composition.
We propose a new approach for spatial verification that aims at modeling object-level regions dynamically clustering keypoints in a 2D Hough space, which are then used to accurately weight small contributing objects within the results, without the need for costly object detection steps.
We call this method 
Objects in Scene to Objects in Scene (OS2OS) score, and it is optimized for fast matrix operations on CPUs.
OS2OS performs comparably to state-of-the-art methods in classic CBIR problems, on the Oxford 5K, Paris 6K, and Google-Landmarks datasets, without the need for bounding boxes.
It also 
succeeds in emerging retrieval tasks such as image composite matching in the NIST MFC2018 dataset and meme-style composite imagery from Reddit.
\end{abstract}

\vspace{-4mm}
\section{Introduction}
\begin{figure}[t]
\begin{center}
\includegraphics[width=1.0\linewidth]{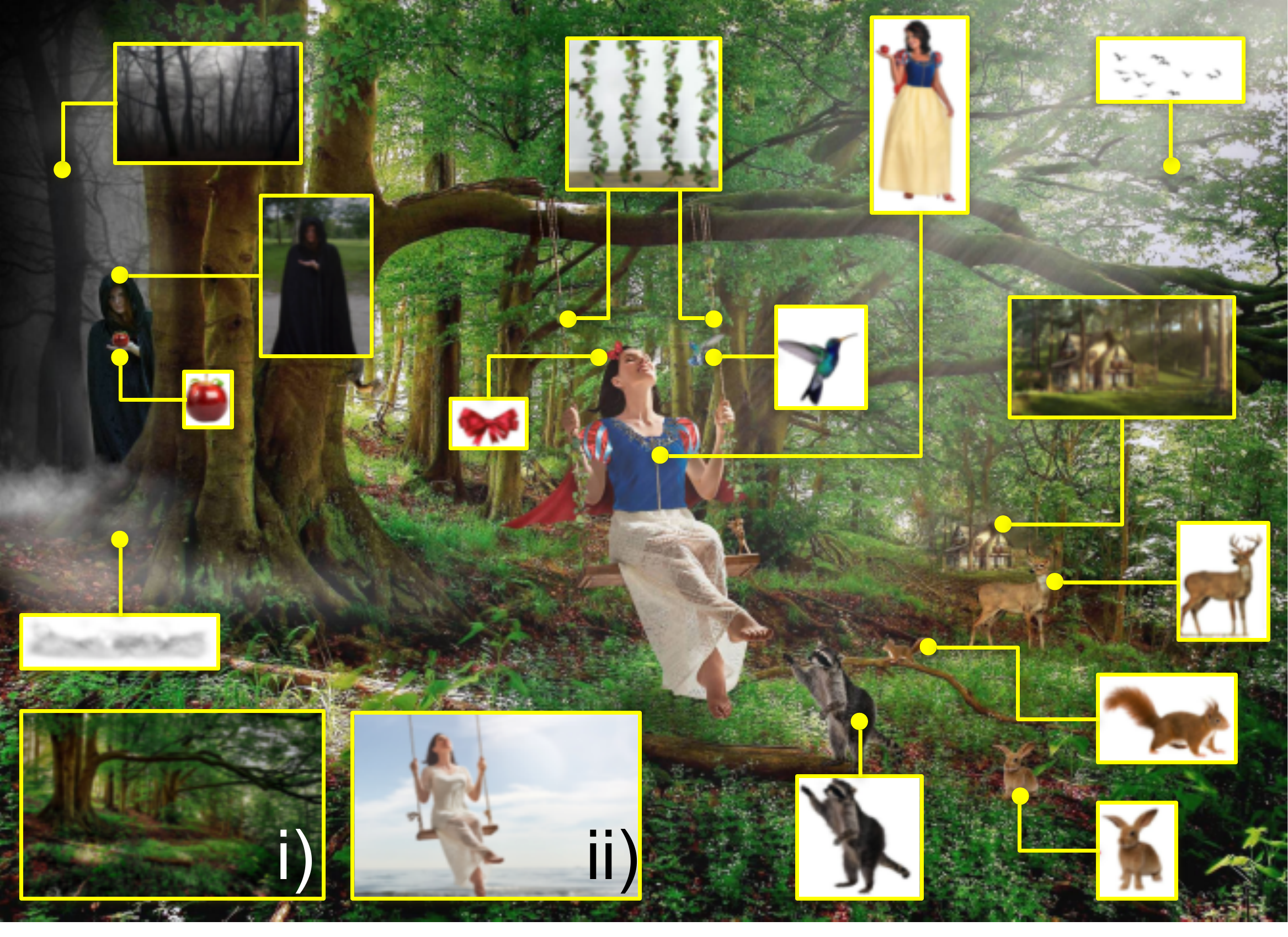}
\end{center}
\vspace{-0.3cm}
\caption{
An example of a meme-style image from the subreddit \url{/r/photoshopbattles}.
Internet memes are humorous messages that are spread on social media, often conforming to a set genre with a distinct style.
In this paper, we provide spatial verification to find object-level correspondences between images, which assist in the retrieval of the donor images (highlighted in yellow) that contribute to composites like the one above.
}
\label{fig:teaser}
\vspace{-0.3cm}
\end{figure}

Image retrieval traditionally starts with the knowledge of what a user is looking for at query time and ends with a set of relevant results that match those initial expectations.
This search regime must be transformed, however, in the age of social media.
Nowadays, complex composite images like Internet memes are pervasive in the pools of content that are shared by users on social media~\cite{baym2015personal}.
Some of these images are of significant cultural value~\cite{shifman2014memes}, while others represent odious extremist propaganda~\cite{bellingcat}.
Both of these cases are important and timely topics of study, where the 
retrieval of the source content in reference image collections is paramount.
The identification of near duplicate content shared between artistic images has received attention recently, and it can be solved using learned features that are both discriminative and invariant~\cite{shen2019discovering}.
Further, the evolution of a meme can be traced by finding all of the related images, including images that contributed small donor objects (Fig.~\ref{fig:teaser}), and referencing associated timestamps~\cite{moreira2018image, bharati2018beyond}.
Similarly, one can debunk misleading or forged images being used for disinformation purposes by verifying identified source material~\cite{brogan2017spotting}.
In this paper we propose a new solution for spatial verification in image retrieval that supports these tasks by modeling object-level regions with the goal of retrieving the sources of small regions in large datasets.

One of the major hurdles of image retrieval is the \emph{semantic gap}~\cite{smeulders2000content, zhou2017recent}, where a user's understanding of the content transcends the low-level representation and subsequent recall capabilities of a content-based search engine.
Ideally, a good algorithm should return results that satisfy the user's higher-level intent, not just images with similar low-level features.
However, retrieval approaches developed around this idea may not be sufficient for parsing composite imagery and retrieving relevant results, especially if there are aspects of an image that are not immediately apparent to the user at query time.
In addition to the context of image creation and use, an image's extent is also shifted as it can be reused in composite images later on.
And finally, given the free-form nature of composite imagery, an effective retrieval approach is expected to operate over a web-scale database to maximize potential matching candidates.

With the above aspects in mind, we aim at adapting image retrieval to different contexts, such as composite images, especially those containing small spliced objects.
For example, in Fig.~\ref{fig:teaser}, we see a composite image created from many smaller donor images.
This type of image poses a significant challenge for existing image retrieval algorithms, because if the host (\textit{i.e.}, background) and donor images are matched globally, the latter group would not be highly scored since the content shared with the composite is very small.
Nonetheless, tasks like meme analysis and disinformation debunking require the retrieval of each meaningful piece of content in an image under scrutiny.
Because the image in question is a conglomeration from many sources, we can assume that the goal of a retrieval system should be to return instances of all images contributing to the composite.

To do so, we propose a new method of spatial verification that allows object-level instance scoring of retrieved results, without the need for costly object detection steps.
We devise a feature-agnostic algorithm that utilizes a geometrically consistent voting measure inspired by 
Hough-based voting techniques, with the major difference of not requiring object regions of interest in the query to be known ahead of time, a manner to make the solution more appropriate to the current reality of unspecified retrieval context.
As we show through experiments, the proposed method quickly and accurately localizes and ranks rigid objects contained within the query image to objects contained within a large image database.
We call this method the Objects in Scene to Objects in Scene (OS2OS) score. 

Through the rest of this paper, we will look at approaches that are related to the proposed OS2OS score, detail our methodology, and then perform experiments utilizing a number of relevant image retrieval datasets and features, including classic handcrafted and contemporary deep feature representations.
In summary, the contributions are:
\vspace{-0.2cm}

\begin{itemize}
\setlength{\itemsep}{-0.1cm}
\item A new perspective on the problem of image retrieval, which aims at addressing the deficiencies in existing problem formulations for retrieval in cases of complex composites and other manipulated images.

\item A new method called OS2OS score for spatial verification of matching objects between images, including very small objects that are important for understanding memes and other emerging Internet media.

\item A series of experiments showing the viability of the approach on the Oxford~5K~\cite{philbin2007object}, Paris~6K~\cite{philbin2008object}, Google-Landmarks~\cite{noh2017large}, and NIST MFC2018~\cite{nist2018dataset} datasets, as well as meme-style imagery from Reddit~\cite{moreira2018image}.

\item A new experimental protocol for the Reddit Photoshop Battles dataset~\cite{reddit2018photoshopbattles}, preparing it to be used for benchmarking potential solutions to retrieve the donors of composite images.
\end{itemize}

\section{Related Work}
The typical CBIR solutions rely on multi-level representation of the images, to reduce the semantic gap between the pixel values and the system user's retrieval intent.
In the lowest levels, typical methods use local features (a.k.a., keypoints) to obtain $n$-dimensional descriptions of the image content, ranging from handcrafted representations, such as SIFT~\cite{lowe2004distinctive} and SURF~\cite{bay2008speeded}, to representations learned via neural networks, such as LIFT~\cite{liu2016fine} and DELF~\cite{noh2017large}.

In the subsequent levels, these local features are then used to index and compare, within the $n$-dimensional space they constitute (a.k.a., the feature space) and through euclidean distance or similar method, pairs of image localities (a.k.a., image patches).
State-of-the-art large-scale indexing solutions comprise methods such as Optimized Product Quantization (OPQ) for approximate nearest neighbor (ANN) search~\cite{ge2013optimized}, with Inverted File Indices (IVF)~\cite{johnson2019billion}.

As proposed by Lowe~\cite{lowe2004distinctive}, two features and their respective image patches are probably a \emph{match} (\textit{i.e.}, they depict the same object or scene region in different configurations), if one is the nearest neighbor of the other within the feature space.
Depending on the nature of the CBIR application, local features may be indexed in ways that use only the feature space and ignore or underutilize the scale, orientation, or $(x, y)$ positions of the features within the images they belong to.
That is the case for bag-of-features~\cite{sivic2003video} and similar approaches~\cite{perronnin2007fisher, jegou2010aggregating}, which are mostly useful for tasks such as retrieving semantically similar images.
In the case of retrieving near-duplicates or finding spliced objects across a dataset, though, these techniques are not enough.
Hence the need for additional spatial verification steps, such as the ones proposed in this work, to ensure that the local features being matched present a geometric coherence in either their scale, orientation, or $(x, y)$ positions.
Local-feature spatial verification methods can be organized into two categories, which are described below.

\textbf{Hypothesis-oriented spatial verification.}
Methods in this category start with a set of spatial transformation hypotheses of one image towards the other (\textit{e.g.}, affine transformation).
As originally proposed in the RANSAC ~\cite{fischler1981random} algorithm, these hypotheses are iteratively generated for random samples of feature matches, and are evaluated according to the overall number of matches that are \emph{inliers} (\textit{i.e.}, matches that comply with the hypothesis).
Aiming to make the process more accurate and deterministic, Philbin et al.~introduced the Fast Spatial Measure (FSM) algorithm~\cite{philbin2007object}, which generates one hypothesis for every single feature match.
Although very accurate, the major drawback of techniques from this category is the large runtime they demand, which is a quadratic function of the number of 
features (as we show through experiments in Fig.~\ref{fig:expTiming}).

\textbf{Hough transform-based spatial verification.}
Methods from this category start with Hough transforms~\cite{duda1972use, ballard1981generalizing} and the computation of histograms for their parameters, where each bin quantifies the number of agreeing feature matches.
Lowe~\cite{lowe2004distinctive} proposed the adoption of four-parameter Hough transforms, computing bins with respect to the product of (i) $x$ and (ii) $y$ 
coordinates, (iii) scale, and (iv) orientation of the features.
The largest histogram bin is then chosen to select a better and potentially reduced set of feature matches, prior to applying RANSAC on them.
Aiming to speed up up the overall process, Avrithis and Tolias later introduced the Hough Pyramid Matching (HPM) strategy~\cite{avrithis2014hough}, which employs a hierarchical voting strategy to recursively split pairs of feature matches into bins in a top-down fashion (from coarser to finer correspondences), as a way to evaluate the pairwise affinities of matches without enumerating all pairs.

Similar to HPM, Li et al.~also suggested the evaluation of pairs of feature matches to develop their Pairwise Geometric Matching (PGM) strategy~\cite{li2015pairwise}.
However, they proposed to use coarser (and therefore faster to verify) two-parameter Hough transforms, relying only on the orientation and scale change between the compared images.
This indeed guarantees a significant speed-up in the spatial verification process, as we show through experiments reported in Fig.~\ref{fig:expTiming}.

Another approach worth mentioning is the one proposed by Shen et al., namely Spatially Constrained Similarity Measure (SCSM)~\cite{shen_SCSM}.
Although this one also makes use of four-parameter Hough transform quantization to enumerate candidate spatial transformations, its difference relies on the method to select the best transformation.
By demanding the establishment of a bounding box over the features of the query object before performing the retrieval task, the algorithm uses the box center to measure the quality of the candidate transformations.
Thus, the best transformation is the one that, after its application, best preserves the spatial relations among the feature positions and the box center.

Lastly, Sch\"{o}nberger et al.~introduced the Vote and Verify (VaV)~\cite{schonberger2016vote} method, where HPM is used to compute votes across the Hough transform hierarchy, prior to verifying transformation hypothesis for only the most voted bins.

\textbf{Putting the proposed method in context with prior work.}
Our method belongs to the latter category of spatial verification techniques and is agnostic to the chosen local features and feature indexing approach.
When compared to the literature, the novelty of this work comprises, besides the unique combination of strategies for the task at hand:

\begin{itemize}
\setlength{\itemsep}{-0.1cm}
\item Inspired by SCSM~\cite{shen_SCSM}, the use of match-set-wise centroids to evaluate the quality of the feature matches, without the need for selecting bounding boxes 
of interest in the query ahead of retrieval time.
These centroids are calculated in a novel way (see Eq.~\ref{eq:centroid}). 

\item Inspired by PGM~\cite{li2015pairwise}, the use of a coarse, fast, but still accurate two-parameter Hough transform quantization, which contrary to PGM, relies on the $x$ and $y$ position coordinates of the matched features, instead of scale and orientation.

\item A novel image retrieval score (OS2OS score, see Eq.~\ref{eq:score}), which is based on two complementary measures of object-level image matching quality (namely, object centrality, and angle coherence), and allows for spatial verification while ranking images.

\end{itemize}

In the following, we detail each step of the proposed solution, as well as provide discussion on the reasons and advantages of adopting each one of the above novel aspects.

\section{Objects in Scene to Objects in Scene (OS2OS) Score}
\label{sec:method}
\begin{figure*}[t]
\begin{center}
\includegraphics[width=1\linewidth]{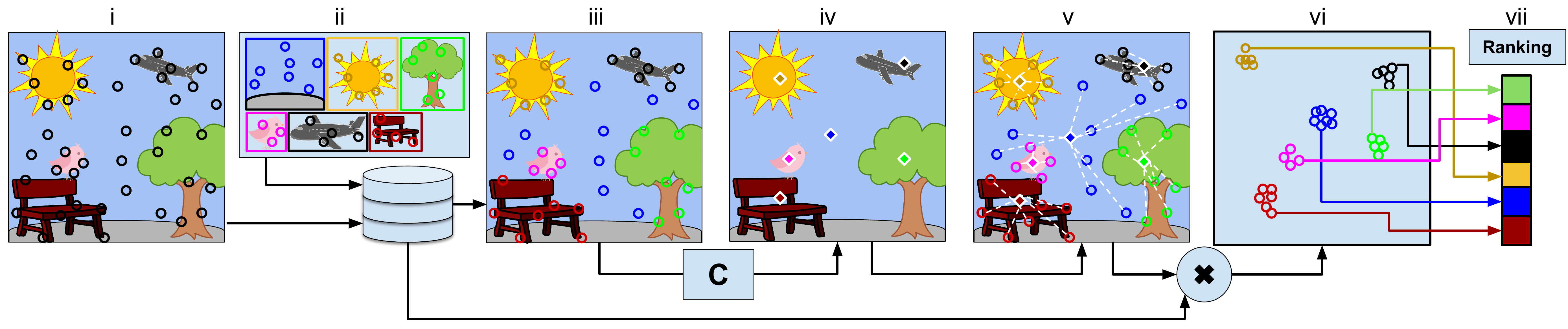}
\end{center}
\caption{
Steps of the OS2OS method.
(i)~Local features with associated geometric data (\textit{i.e.}, coordinates, scale, and rotation) are extracted from the query.
(ii)~An image database is collected that contains donor objects.
(iii)~Query features are assigned to corresponding database matches (represented by feature colors).
(iv)~For each database image sharing matches with the query, a feature centroid is computed, considering only the matched features.
(v)~Keypoint geometric transformations are calculated relative to the estimated centroids.
(vi)~Geometric transformations are applied to the database features, which are clustered in the query $(x, y)$ space.
(vii)~As each cluster represents a potentially shared object, image ranking scores are calculated on an object-by-object basis.
}
\label{fig:pipeline}
\end{figure*}

Our proposed spatial verification method for image retrieval can be explained in five steps.

\vspace{0.15cm}
\textbf{Step 1: Local feature affinity.}
Images are described through local feature vectors and their respective geometric data, namely $(x, y)$ location, scale, and orientation angle.
Let $Q$ be the set of all features extracted from the query, as depicted in Fig.~\ref{fig:pipeline} (i), and $D$ be the set of all features extracted from a target image database, as depicted in Fig.~\ref{fig:pipeline} (ii).
For a particular query feature $q_i \in Q$, with $i = \{1 \ldots |Q|\}$, we compute the $K$ nearest neighbors $d_j \in D$ in the feature space only, ignoring the images they come from.
As a result, $q_i$ participates in $K$ matches $m_{{ij}}=(q_i,d_j)$, where $j=\{1 \ldots K\}$.
A score $S$ for a given match $m_{{ij}}$ that can express the affinity between $q_i$ and $d_j$ is calculated via the $L_2$-distance and the rank-adaptive scoring outlined in~\cite{jegou2011exploiting}:
\begin{equation}
\label{eq:rankadtscore}
S(m_{{ij}}) =  {max}(0, \left \| q_i, d_\phi  \right \|_2 - \left \| q_i, d_j \right \|_2),
\end{equation}
where $\phi$ is a fixed rank position of reference (usually $K/2$).

\vspace{0.15cm}
\textbf{Step 2: Image-pairwise centroid calculation.}
Take an image $P$ from the database that shares content with the query.
There might be a set of feature matches between them, whose incident locations $(x,y)$ onto the query should give a rough indication of the shared object's location within $Q$.
Therefore, if we calculate the center of these $P$-wise match locations on the query, we find a point that is \textit{generally} near a potential object of interest, serving as an estimation of its centroid $c$ (see Fig.~\ref{fig:pipeline} (iv)).
Let $M$ be the set of feature matches $m_k$ shared between $P$ and the query, with $k=\{1 \ldots |M|\}$, and let $Q_m \subseteq Q$ be the set that contains only the query features that have a match to $P$.
To obtain $c$, we apply the straightforward solution of using the Euclidean center of the query features $q_k \in Q_m$.
Nevertheless, aiming to consider the quality of the matches while computing $c$, and to deal with spurious matches, we also weight the added features according to their affinity scores $S(m_k)$ (see Eq.~\ref{eq:rankadtscore}) associated with the features of $P$:
\begin{equation}
\label{eq:centroid}
c=\frac{\sum_{k=1}^{|M|}L(q_k)\times S(m_k)}{\sum_{k=1}^{|M|}S(m_k)},
\end{equation}
where $L(q_k)$ is the $(x,y)$ location of the $k$-th feature $q_k \in Q_m$, and $m_k \in M$ is the respective $k$-th feature match between the query and $P$.
The motivation for this is that the more similar two matched features are (\textit{i.e.}, the higher their value of $S(.)$), the more they contribute to the position of $c$.
This strategy reduces the Hough voting noise problem described in~\cite{schonberger2016vote}, as shown via supplemental ablation experiments (see Supp.~Mat.), where we report the decrease in performance due to the absence of centroid computation.

\begin{figure*}[t]
\begin{center}
\includegraphics[width=1.0\linewidth]{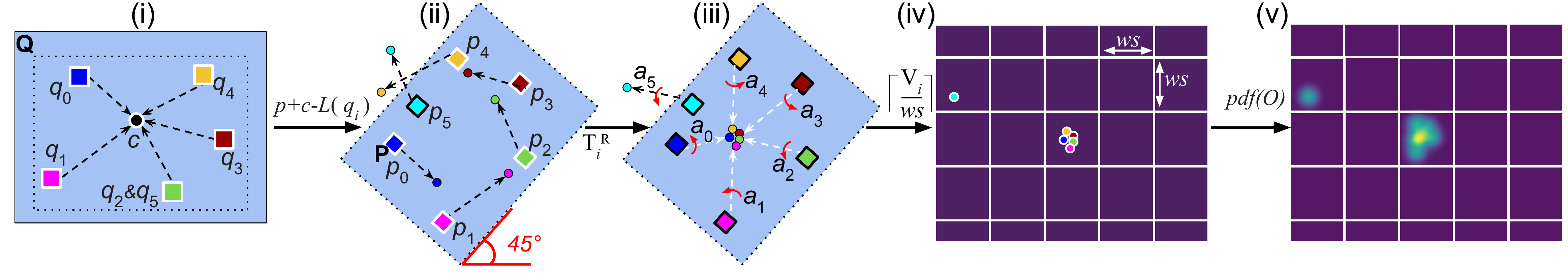}
\end{center}
\caption{
OS2OS voting space computation.
(i)~Translation vectors are calculated relative to a computed centroid $c$ in $Q$.
(ii)~Translations are applied to matching features in $P$.
(iii)~Vectors are rotated by the angle differences $a_k$ to account for object rotation.
Notice that the $p_5$ spurious match votes incorrectly and therefore will not contribute to the score.
(iv)~Votes are binned using $(x, y)$ coordinates only.
(v)~A density map is calculated, which conveniently coincides with the query space, providing visualization of matched object regions.
}
\label{fig:voting}
\end{figure*}

\vspace{0.15cm}
\textbf{Step 3: Centroid-relative feature projection.}
Given the centroid $c$ representing the query feature locations $q_k \in Q_m$, and their respective matched features $p_k \in P$, we can estimate the translation, rotation and scaling transformation from the space of image $P$ towards the query space, for each match $m_k=(q_k, p_k)$, with $k = \{1 \ldots |M|\}$: 
\begin{equation}
\label{eq:totalT}
T_k = T^R_k\cdot(c-L(q_k))\times\frac{\sigma(p_k)}{\sigma(q_k)},
\end{equation}
\begin{equation}
\label{eq:rotT}
T^R_k = \begin{bmatrix}
cos(a_k) & -sin(a_k)\\
sin(a_k) &  cos(a_k)
\end{bmatrix}, a_k=\theta(p_k)-\theta(q_k)
\end{equation}
where $\theta(.)$ and $\sigma(.)$ respectively provide the angle and the scale associated with the location $L(.)$ of either $q_k$ or $p_k$ features.
These transformations describe where each query feature expects the object region to be, similar to an ``R-table'' of the general Hough transform~\cite{ballard1981generalizing}, with the novelty that we only consider the $(x,y)$ feature coordinates.
The advantage of doing so is twofold: (i) the Hough space is two-dimensional (given $x$ and $y$) instead of four-dimensional, making computations faster; (ii) the Hough space maps directly to the query space, making the localization of shared objects straightforward.
By applying each transformation to the $(x,y)$ location of its respective matched feature $L(p_k)$, we subsequently build a voting space $V_k$ for each $p_k \in P$:
\begin{equation}
\label{eq:vote}
V_k = L(p_k) + T_k.
\end{equation}
The process of computing the voting space is shown in Fig.~\ref{fig:voting}.
Observe, through item (iii), that all $p_k$ features contribute to the centroid, except for the spurious $p_5$ one.

\vspace{0.15cm}    
\textbf{Step 4: Density-based feature clustering.}
We calculate a distinct centroid $c$ for every image $P$ from the database that presents matches with the query.
This allows us to transform all matched feature vectors, regardless of the image they come from.
After all matched feature locations have been transformed to their respective two-dimensional Hough vote space, as depicted in Fig.~\ref{fig:pipeline} (vi), a density-based clustering determines which sets of feature matches are structurally consistent with the query.
Instead of using a computationally intensive algorithm such as DBSCAN~\cite{ester1996density} for clustering millions of points, we apply a linear-time grid-based quantization to each $V_k$, to find the approximate clusters of highest vote density.
Because we assume, for simplicity sake, cluster morphologies to be roughly circular, we employ a square sliding window to quantize and bin $V_k$ values.
The size $ws$ of the quantization window varies according to the resolution $w \times h$ of image $P$:
\begin{equation}
\label{eq:ws}
ws = \left ( \frac{max(w, h)}{b} \right)^{1-\epsilon},
\end{equation}
where $b$ is a scaling factor, and $\epsilon \asymp 0$ prevents $ws$ from becoming too large.
The magnitudes of projection vectors $||T_k||$ are proportional to the resolution of image $P$.
As $||T_k||$ increases, small errors in the scale normalization and rotation transforms (Eq.~\ref{eq:rotT}) are amplified, resulting in lower density clustering.
Unlike HPM~\cite{avrithis2014hough}, which accounts for this phenomenon by utilizing a hierarchy of window sizes for assigning votes to clusters, we employ a single window size determined by a sub-linear mapping of the image's resolution (Eq.~\ref{eq:ws}).
This is accomplished through $\epsilon$: it constrains the maximum allowable vote cluster area to grow sub-linearly with the resolution of the image.
The values of $b$ and $\epsilon$ are empirically determined via an ablation study, included in the Supp.~Mat. 

\vspace{0.15cm}
\textbf{Step 5: OS2OS filtering and scoring.}
By relying on the value of $ws$ computed for an image $P$, the respective votes $V_k$ are quantized into cluster bins (see Fig.~\ref{fig:voting}~(iv)):
\begin{equation}
\label{eq:bin}
bin(V_k) = \ceil[\Big]{\frac{V_k}{ws}}.
\end{equation}
We consider that all $V_k$ values sharing the same $bin(.)$ value belong to the same matched object.
Let $O$ be the set of transformed features from database image $P$, which participate in matches between $P$ and the query, and whose respective votes happened to belong to the same bin according to Eq.~\ref{eq:ws}.
The meaning of this is that the $O$ features likely belong to a unique object shared by the images. Subsequently, we enforce a strict one-to-one matching constraint within each $O$. Then,
to express their affinity of features within the filtered $O$, we propose two novel main scoring mechanisms inspired by \cite{wu2015robust}, each with a particular purpose.

The first, called the Centrality Score ($CS$), measures the centrality of the features and is calculated as the sum of location differences between the elements $o_k \in O$ and their average (a.k.a., central) element $\bar{o}$, with $k = \{1 \ldots |O|\}$:
\begin{equation}
\label{eq:cs}
CS = \frac{\sum_{k=1}^{|O|}pdf(\left \| L(o_k)-L(\bar{o}) \right \|_2)}{|O|},
\end{equation}
where $L(.)$ is, again, the $(x,y)$ location of the given feature, and $pdf(.)$ is a function that augments the available data through a probability density function (see Fig.~\ref{fig:voting}~(v)): 
\begin{equation}
\label{eq:pdf}
pdf(x) = \frac{e^{-x^2/2}}{\sqrt{2\pi}}.
\end{equation}
In addition, $CS$ is normalized by the cardinality of $O$, as a way to balance clusters of small objects containing few features with respect to large objects that contain many.

The second mechanism is the Angle coherence Score ($AS$), which aims at measuring the uniformity of angles within the features of $O$.
Consider the feature-wise difference of angles $a_k$ expressed in Eq.~\ref{eq:rotT}, and let $A$ be the set of difference of angles $a_k$ computed for each feature of $O$.
Features from keypoints belonging to a single rigid object are expected to present similar values of $a_k$, while erroneously aggregated unrelated features are expected to present more diverse results.
For that reason, we rely on the inverse of the standard deviation of $A$, $stdv(A)$, to compute the angle uniformity (shifted to avoid division by zero):
\begin{equation}
\label{eq:as}
AS = 1 / (1 + stdv(A)).
\end{equation}

Finally, the OS2OS score is given by $CS$ and $AS$:
\begin{equation}
\label{eq:score}
OS2OS = CS \times AS \times \log{|O|},
\end{equation}
which lays between $0$ and $1$ and rewards only clusters in which each point has been transformed similarly.
The logarithmic scalar $\log{|O|}$ penalizes clusters with very low numbers of votes, pushing their score towards zero.

\section{OS2OS Score Evaluation}
\label{sec:Eval}
\textbf{Datasets.}
\textit{Oxford~5k and Paris~6k.}
Oxford~5k~\cite{philbin2007object} and Paris~6k~\cite{philbin2008lost} are smaller popular datasets for image retrieval performance evaluation.
These two datasets contain hundreds of true positive matches per query, rather than similar sized datasets that contain only four or five~\cite{nister2006scalable, jegou2008hamming}.

\textit{Google-Landmarks.}
\label{GoogleDS}
As a benchmark for DELF features~\cite{noh2017large}, Google released the 2018 Google-Landmarks dataset~\cite{GLandMarks} and subsequent evaluation protocol~\cite{GLandMarksChallenge}.
This dataset contains $1,098,461$ images with $117,703$ queries in the testing protocol.
While groundtruth data for the test set has not yet been released, $1,212,281$ weak labels for the training set are available.
The training set contains a total of $14,951$ unique landmarks, with each landmark having an average of $80$ instances.

\textit{NIST Media Forensics Challenge 2018 (MFC2018).}
As part of the yearly \textit{Media Forensics Challenge} run by NIST, the MFC2018 dataset~\cite{nist2018dataset} was constructed specifically for the task of finding related manipulated images in a forensics context.
This is a large dataset, $3.1$TB in size, containing $1,031,080$ images and $3,300$ queries.
Many of these queries are composites, with groundtruth results provided in the MFC2018 testing protocol.
The dataset also provides groundtruth as to whether a database image contributes a majority of its content (\textit{e.g.}, Fig.~\ref{fig:teaser} i), or only a particular object (\textit{e.g.}, Fig.~\ref{fig:teaser} ii) to the query.

\textit{Reddit.}
\label{RedditDS}
The Reddit dataset~\cite{redditDataset} introduced by Moreira et al.~\cite{moreira2018image} is collected from the Reddit Photoshop Battles~\cite{reddit2018photoshopbattles} subreddit.
Each case provides the original image and all subsequent manipulated versions of the original.
This dataset contains $51,245$ images from $185$ different Photoshop battles.
To utilize this dataset for the OS2OS task, we generate a query set of one image chosen at random from each of the $185$ cases.
The rest of the images from the same battle thread are considered as the groundtruth for relevant images to these queries.
We will make this new groundtruth available upon the publication of this paper.


\begin{table}[t]
\renewcommand{\arraystretch}{1.1}
\centering
\begin{tabular}{L{1.6cm}R{1.7cm}R{1.9cm}R{1.4cm}}
\hline
Local Feature & Dimensions (\#) & Features per~image~(\#) & Time (sec.) \\
\hline
DSURF     & 64   & 1000 & 0.06 \\
DELF      & 40   & 1000 & 60.04 \\
LIFT      & 64   & 500  & 182.22 \\
MobileNet & 1280 & 1    & 1.09 \\
\hline
\end{tabular}
\vspace{0.1cm}
\caption{
Computation time per image for a subset of the \textit{Reddit} dataset.
Features for $100$ images were calculated, and each image was 1MB in size on average.
}
\label{tab:exTimes}
\vspace{-0.2cm}
\end{table}

\textbf{Features.} For each dataset, we report performance using both handcrafted SURF~\cite{bay2008speeded} and learned DELF~\cite{noh2017large} features.
While other deep image representations have been proposed for image retrieval~\cite{howard2017mobilenets, yi2016lift,gordo2016deep,radenovic2016cnn}, 
some deep local feature descriptors such as LIFT~\cite{liu2016fine} are too slow (see Table~\ref{tab:exTimes}) for practical use, as also observed in~\cite{noh2017large}.
Other fast global descriptors such as MobileNet~\cite{howard2017mobilenets} are not applicable to the localization of multiple objects.
Our region-wise matching approach requires local descriptors to match coherently within a certain spatial location, which cannot happen with a single global MobileNet descriptor.

SURF keypoints are detected in a distributed modality (DSURF), as proposed in~\cite{moreira2018image}.
Keypoint extraction of DSURF features automatically provides the location, scale, and rotation data needed for computing the OS2OS score.
For all datasets, we extract a maximum of $5,000$ $64$-dimensional DSURF features per image, along with their respective geometrical data. 
DELF features are used in similar way to DSURF.
Because the DELF algorithm provides only feature scale information, we use the SURF keypoint angle algorithm~\cite{bay2008speeded} as an extension to provide feature angles for the DELF geometric data.
We performed experiments using the default parameters and model to produce 40-dimensional local features, and cap the attention model at a maximum of $1,000$ features per image.
 

\begin{figure}[t]
\vspace{-0.1cm}
\begin{center}
\includegraphics[width=0.9\linewidth]{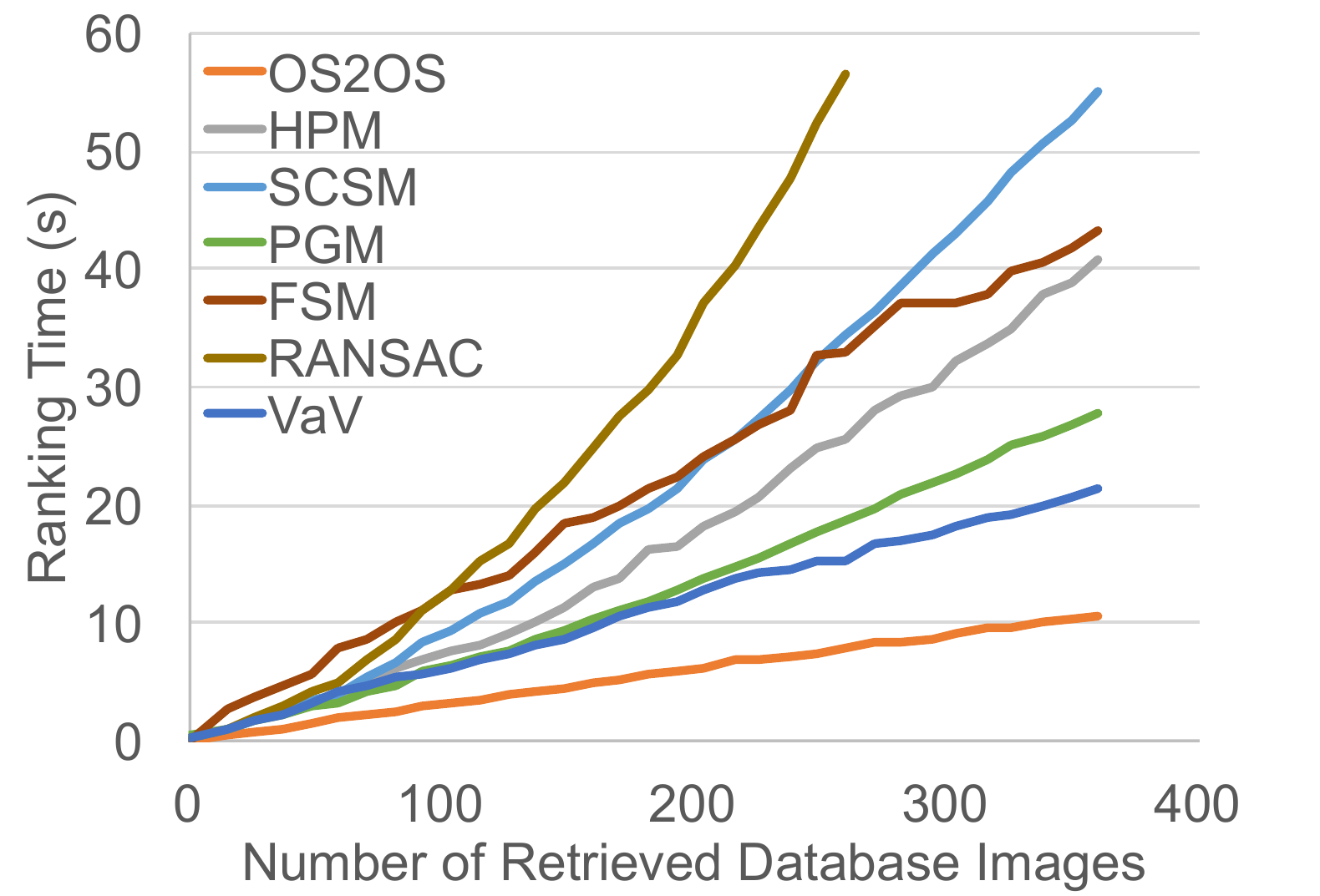}
\end{center}
\vspace{-0.2cm}
\caption{
Spatial verification (ranking) timings for different algorithms.
Each algorithm is given an identical subset of $K$ nearest-neighbor features for a query.
The lower the ranking time, the better its performance.
Results are for the \textit{Google-Landmarks} dataset.
}
\label{fig:expTiming}
\vspace{-0.2cm}
\end{figure}

\textbf{Indexing.}
For experiments on \textit{Oxford~5k} and \textit{Paris~6k} datasets, we keep a consistent indexing backbone among spatial 
verification methods.
To index features, we use Optimized Product Quantization (OPQ) for Approximate Nearest Neighbor (ANN) search, with Inverted File Indices and Asymmetric Distance Computation (IVFADC)~\cite{johnson2019billion}.
Following the vocabulary hold-out protocol suggested in~\cite{tolias2013aggregate}, we utilize a randomized $1\%$ subset of \textit{Google Landmark} images to train the quantization table used for \textit{Oxford~5k}, \textit{Paris~6k}, and \textit{Reddit} datasets.
Due to the large volume of images available, a randomized $5\%$ hold-out of the \textit{Google Landmarks} dataset is used to train its own tables.
These tables are obtained via OPQ matrix computation~\cite{ge2013optimized}, and used for the IVFADC centroids.
OPQ training runs for $25$ epochs for $2^{18}$ centroids using four NVIDIA TITAN Xp GPUs. 
This process results in a total of four IVFADC tables: two for \textit{Oxford~5k}, \textit{Paris~6k}, and \textit{Reddit} datasets (trained on either DSURF or DELF features),
and two for \textit{Google Landmarks} (trained on either DSURF or DELF).
The resulting IVFADC structures are used to index all local image features from their respective datasets.



\section{Experimental Results}
\vspace{-0.1cm}
Here we describe and analyze the results for a series of timing experiments, as well as the experiments for  each of the five datasets described in Sec.~\ref{sec:Eval}. 

\subsection{Timing and Complexity}
\textbf{Feature extraction timings.} Aiming to focus on large-scale retrieval, we performed average timing experiments over a subset of $100$ images from the \textit{Reddit} dataset to analyze the tractability of different feature extraction methods on CPUs.
Table~\ref{tab:exTimes} shows these results, justifying our
selection of DSURF and DELF as candidate features for further experiments.
MobileNet stands for the MobileNet-v2 architecture~\cite{howard2017mobilenets}, whose features were extracted from its \textit{global\_pool} layer.
Although fast, local spatial information is not native to MobileNet, and is therefore incompatible with the task of spatial verification.

\begin{table}[t]
\renewcommand{\arraystretch}{1}
\small
\centering
\begin{tabular}{L{1.9cm}R{1.1cm}R{1cm}R{1.1cm}R{1cm}}
\hline
 & \multicolumn{2}{c}{Oxford~5k} & \multicolumn{2}{c}{Paris~6k} \\
\cmidrule(lr{1em}){2-3}
\cmidrule(lr{1em}){4-5}
 & DSURF & DELF & DSURF & DELF \\
\hline
Feature-Only$^\dagger$  & 66.7 & 81.5 & 63.8 & 83.6 \\
HPM$^\dagger$  & 72.5 & 82.2 & 69.3 & 83.6 \\
PGM$^\dagger$  & 75.2 & 81.9 & 73.5 & 83.8 \\
VaV$^\dagger$  & 77.4 & \textbf{83.6} & \textbf{74.6} & 81.2 \\
\textbf{OS2OS (ours)} & \textbf{77.9} & 83.1 & 74.1 &  \textbf{86.7} \\
\hline
\end{tabular} \\
$^\dagger$\footnotesize Uses groundtruth bounding boxes to pre-select regions of interest.
\vspace{0.1cm}
\caption{
Mean Average Precision (MAP) scores of different algorithms for queries in the \textit{Oxford ~5k} and \textit{Paris~6k} datasets.
OS2OS scoring provides competitive or superior performance, without the need for bounding boxes to pre-select regions of interest.
}
\label{tab:oxpar}
\vspace{-0.2cm}
\end{table}

\textbf{Spatial Verification Timings.}
We also performed timing experiments against other spatial verification and ranking methods for image retrieval.
Here we compared the OS2OS score against HPM~\cite{avrithis2014hough}, PGM~\cite{li2015pairwise}, FSM~\cite{philbin2007object}, SCSM~\cite{shen_SCSM}, plain RANSAC~\cite{fischler1981random}, and VaV~\cite{schonberger2016vote}.
We extracted $5,000$ SURF features from a query image and varied the $K$ retrieved nearest neighbors for each query feature from $0$ to $400$.
For these experiments we averaged timings across $10$ query images from the \textit{Google-Landmarks} dataset.
Standard deviations were calculated, but were too small to be plotted.
While RANSAC is known to provide spatial verification for an arbitrary number of features in quadratic time~\cite{li2015pairwise}, HPM and PGM are proven to be linear~\cite{avrithis2014hough, li2015pairwise}.
As can be seen in Fig.~\ref{fig:expTiming}, OS2OS scoring is faster than HPM, PGM, and VaV, ranking nearly 400 images with $1.8$ million features in only $10$ seconds.
All methods used the same 2.7GHz single-core environment.



\subsection{Image Retrieval}
\textbf{Oxford~5k and Paris~6k.}
The small-scale experiments performed on the \textit{Oxford~5k} and \textit{Paris~6k} datasets are meant to show that the OS2OS scoring algorithm, while designed for object-level spatial verification, is general enough to provide benefits for typical image retrieval.
We compare our approach against HPM~\cite{avrithis2014hough}, PGM~\cite{li2015pairwise}, VaV~\cite{schonberger2016vote}, and plain usage of SURF and DELF features (without spatial verification).
Mean Average Precision (MAP) scores are reported in Table~\ref{tab:oxpar}.
We find that the OS2OS score significantly improves both DSURF and DELF features, suggesting that the provided spatial constraints work satisfactory for global geometric verification for instance retrieval.
Additionally, we see a much larger performance improvement for DSURF, which suggests the OS2OS score helps mitigate erroneous matches from the bursty nature of SURF.
Overall, the OS2OS approach is comparable to other approaches in the literature.
Of noteworthy significance is the fact that the OS2OS algorithm required no bounding boxes, still performing comparably to other spatial methods.
We additionally examined scores from RANSAC~\cite{fischler1981random}, SCSM~\cite{shen_SCSM}, and FSM~\cite{philbin2007object} for \textit{Oxford~5k}, but the results were inferior to the better performing approaches in Table~\ref{tab:oxpar}.

\begin{figure}[t]
\begin{center}
\includegraphics[width=1\linewidth]{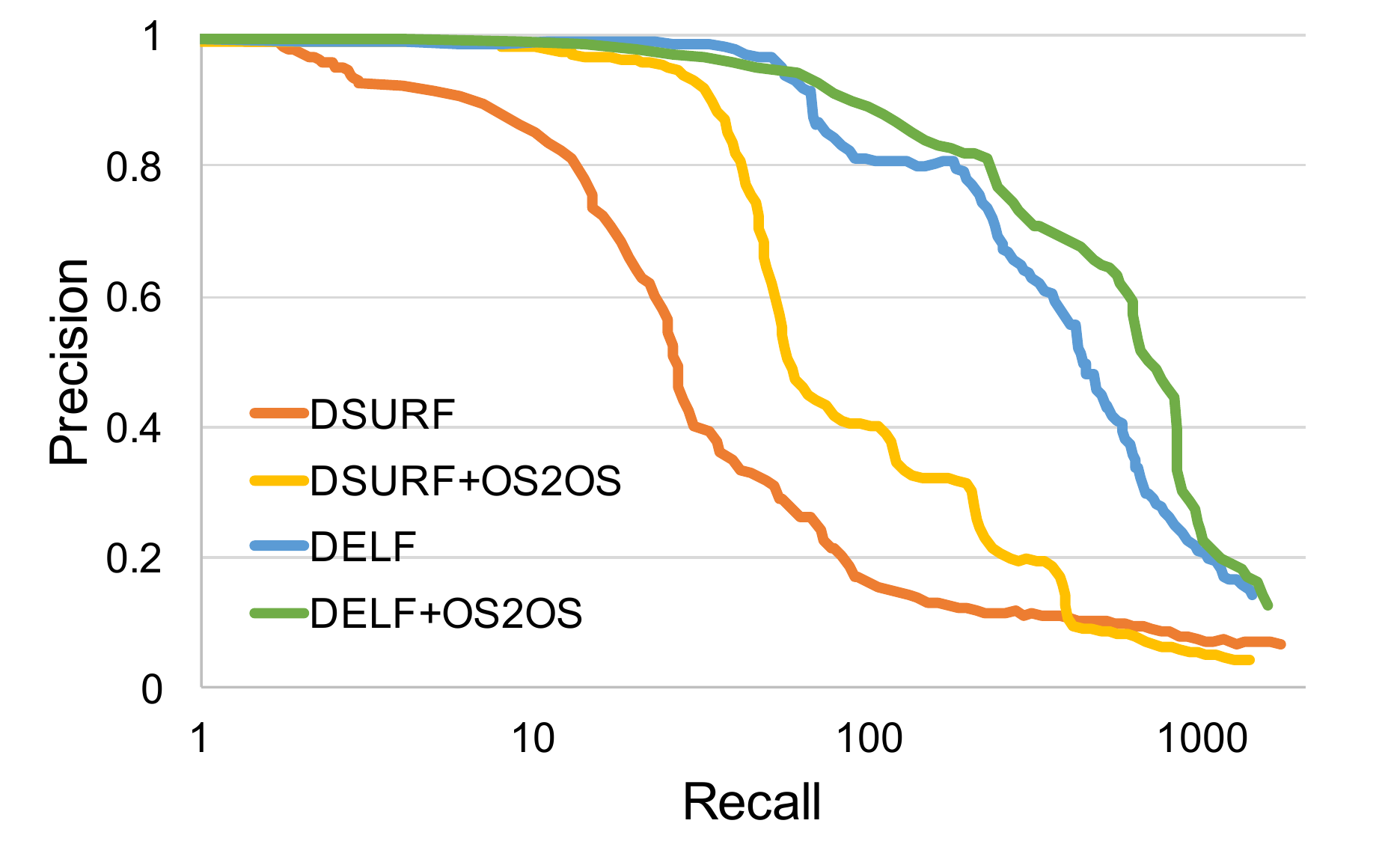}
\end{center}
\vspace{-0.4cm}
\caption{
Precision-recall curves for the \textit{Google-Landmarks} dataset. 
As it can be observed, the usage of spatial verification through the OS2OS score boosts both SURF- and DELF-feature-based image instance retrieval.
}
\label{fig:expGLD}
\end{figure}

\begin{figure}[t]
\begin{center}
\includegraphics[width=1\linewidth]{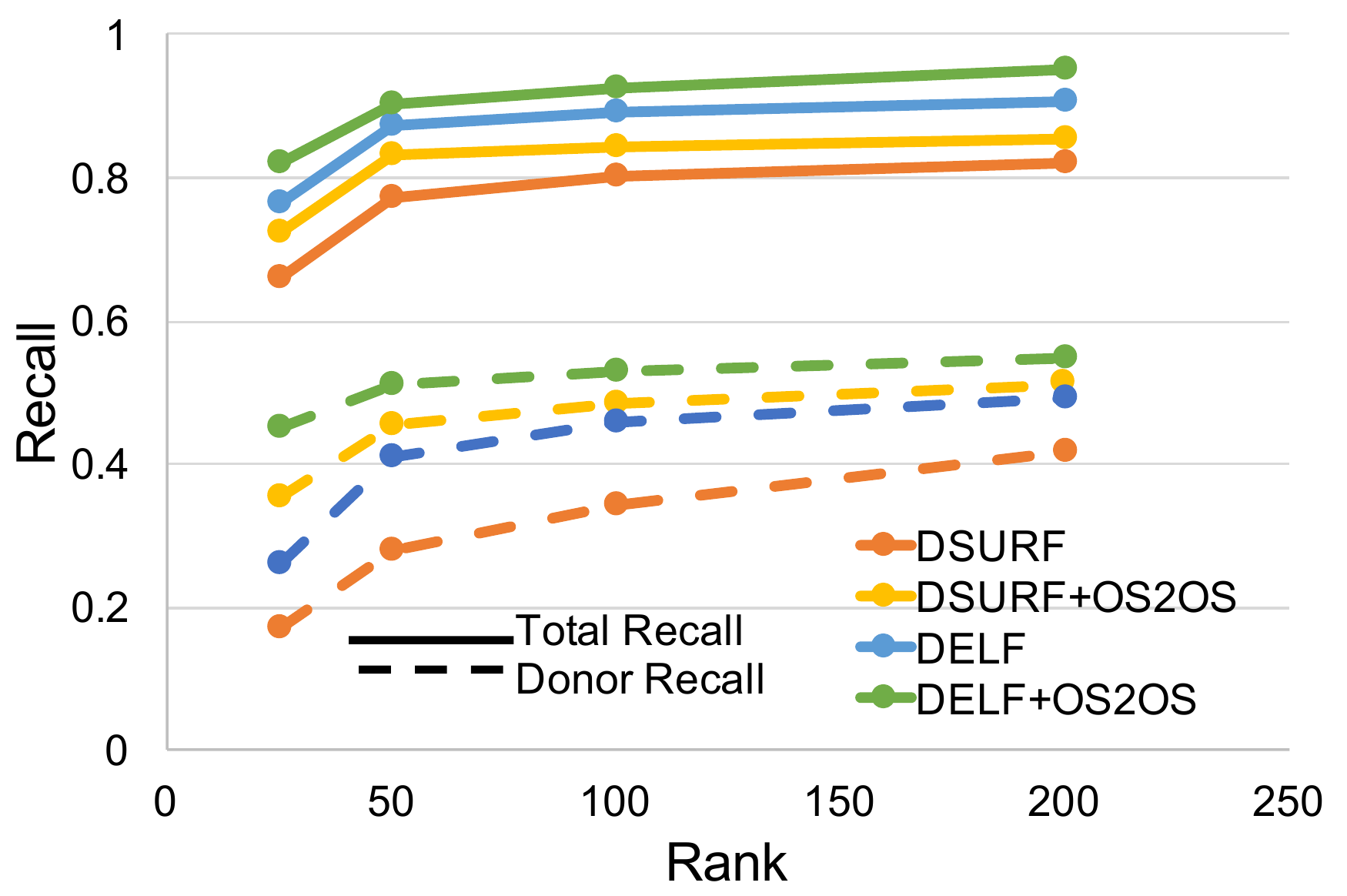}
\end{center}
\vspace{-0.2cm}
\caption{
Recall scores for the \textit{NIST MFC2018} dataset for ranks of 25, 50, 100, and 200 images.
Total recall is represented by solid lines, while small-donor-only recall is represented by dashed lines. OS2OS scoring improves retrieval in all scenarios.
}
\label{fig:expNimble}
\end{figure}

\textbf{Google-Landmarks.}
This experiment shows that the OS2OS score generalizes to the instance retrieval task, as well as showcases the algorithm's scalability in the presence of many distractors.
Because annotations for the index or test sets have not been released at the time of this writing, we performed our study utilizing the training set, which contains over 1 million images and provides landmark annotations.
We selected $1,000$ landmarks at random for retrieval, sampling one query per landmark. Finally, we utilized the modified precision and recall measures described in~\cite{noh2017large} to report results.
Fig.~\ref{fig:expGLD} shows that DSURF augmented with the OS2OS score improves significantly, while also providing minor improvements to DELF.

\textbf{NIST MFC2018.}
Unlike the experiments described thus far, the \textit{MFC2018} dataset is comprised of manipulated images.
Query images from the dataset's retrieval protocol may or may not contain regions from multiple sources within the image database.
We utilized the groundtruth relationship graphs to determine which images donate small objects to their respective queries.
Using this data we can generate recall curves exclusively for donor image retrieval (namely, donor recall).
We report both total recall scores and donor recall scores in Fig.~\ref{fig:expNimble}.
While the OS2OS score shows good boosts in total recall, we see that donor recall is more significantly improved, indicating that the OS2OS score is capable of balancing geometrically coherent matches of image regions from small donors with global matches from backgrounds.

\begin{figure}[t]
\begin{center}
\includegraphics[width=1\linewidth]{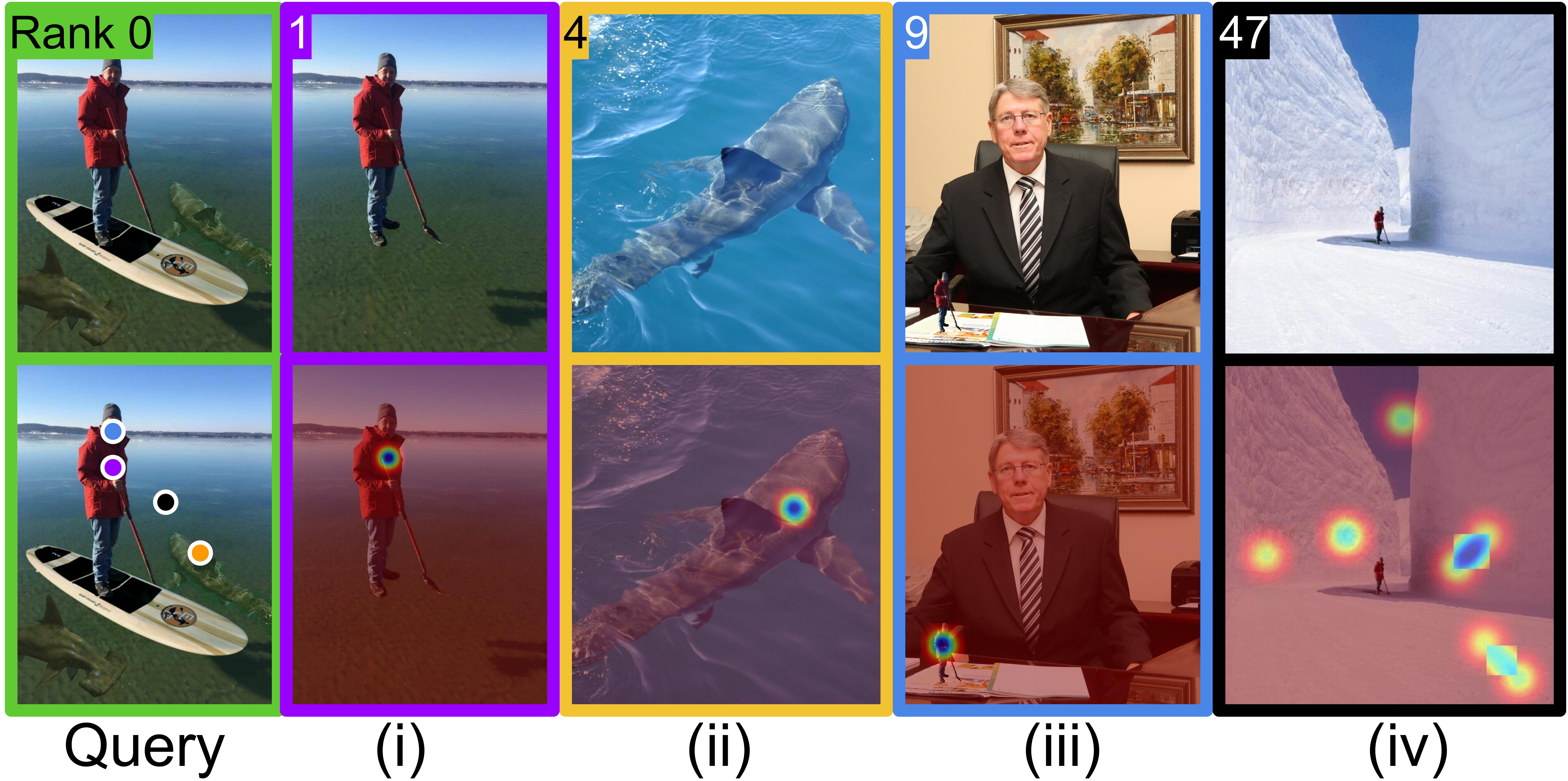}
\end{center}
\caption{
Set of retrieved images for a query (left) in the \textit{Reddit} dataset.
Top row shows retrieved results.
Bottom row provides an overlay of the feature vote space $pdf(V_k)$ from Eq.~\ref{eq:vote}.
Object centroids for each retrieved image are color-coded and overlayed on the query.
Rank 1 (i) is the unmodified version of the query.
Rank 4 (ii) is an object donor to the query, while rank 9 (iii) utilizes a scaled version of the man in the query.
Rank 47 (iv) is a failure case.
See Supp.~Mat.~for additional examples.
}
\label{fig:qual_reddit}
\end{figure}

\textbf{Reddit.}
The \textit{Reddit} provenance dataset has proven to be a difficult challenge~\cite{moreira2018image}.
In Table~\ref{tab:reddit}, we see a significant increase in retrieval performance (nearly $10\%$ for SURF and nearly $5\%$ for DELF) across the board, with vastly superior results when compared to VaV~\cite{schonberger2016vote}.
The near-baseline performance of VaV suggests that global spatial verification methods are not fully adequate to solve the OS2OS problem. 
Further, Fig.~\ref{fig:qual_reddit} shows qualitative retrieval results using OS2OS scoring, along with the object vote maps from each retrieved match.
These results highlight the OS2OS score's ability to localize and appropriately weight small objects from a database image to small objects within the query, without the need for bounding boxes.

\begin{table}[t]
\renewcommand{\arraystretch}{1.1}
\centering
\begin{tabular}{L{2.6cm}R{1.3cm}R{1.3cm}R{1.3cm}}
\hline
Method & R@50 & R@100 & R@200 \\ 
\hline
DSURF         &          0.317 &         0.432  &         0.478 \\
DELF          &          0.402 &         0.516  &         0.551 \\
DSURF + VaV   &          0.310 &         0.423  &         0.479 \\
DSURF + OS2OS &          0.424 &         0.509  &         0.546 \\
DELF + OS2OS  & \textbf{0.479} & \textbf{0.548} & \textbf{0.593} \\
\hline
\end{tabular}
\vspace{0.1cm}
\caption{
Recall scores for the \textit{Reddit} dataset at the top-50, 100, and 200 most related retrieved images. OS2OS spatial verification improves the results in all scenarios.
}
\label{tab:reddit}
\end{table}

\subsection{Parameter Ablation}
To examine the contribution that different parameters from Sec.~\ref{sec:method} provide for both total and donor recall metrics (see Fig.~\ref{fig:expNimble}), we perform an ablation study using the \textit{MFC2018} dataset.
We remove and re-introduce the $CS$ score (Eq.~\ref{eq:cs}), $AS$ score (Eq.~\ref{eq:as}), logarithmic scaling (Eq.~\ref{eq:score}), and centroid calculation (Eq.~\ref{eq:centroid}), to empirically show that each component plays an important role in increasing the performance of the algorithm.
As a consequence, the best configuration is indeed the one that uses all of these features.
We also vary the bin size $ws$ in Eq.~\ref{eq:ws}, to illustrate its effect on filtering and scoring performance.
These additional results are included in the Supp.~Mat., due to space constraints.

\section{Conclusions}
Retrieving images that share small regions in a complex composite scene is a challenge for most image retrieval approaches. In this paper, 
we proposed an inexpensive scoring technique that is based on the better utilization of pre-trained feature extractors and indexing techniques to yield geometrically consistent localized scores. An advantage of the technique is that it is learning-free, and works as an add-on to existing feature description methods. It provides a way to include spatial verification while performing matching in a large feature space. It is optimized to be efficiently executed on CPUs without the need for high-end GPUs. 



In experiments, the proposed approach
improved recall for the retrieval of images for difficult emerging problems such as cultural analytics and image forensics.
As with most computer vision algorithms, the proposed approach still struggles on challenging datasets obtained from the web with different styles of content correspondence.
Although performance improves with the proposed scoring, retrieval for specific applications such as tracking memes on social media 
is worthy of additional research consideration --- especially as such content grows in popularity.



{\small
\bibliographystyle{ieee}
\bibliography{egbib}
}

\end{document}


\title{Dynamic Spatial Verification for Large-Scale Object-Level Image Retrieval}


\maketitle

\section{Optimizing the NH Score for CPU and GPU computation}
\begin{figure*}[htb]
\hspace*{-1.3cm}  
\centering
   \includegraphics[width=1.2\linewidth]{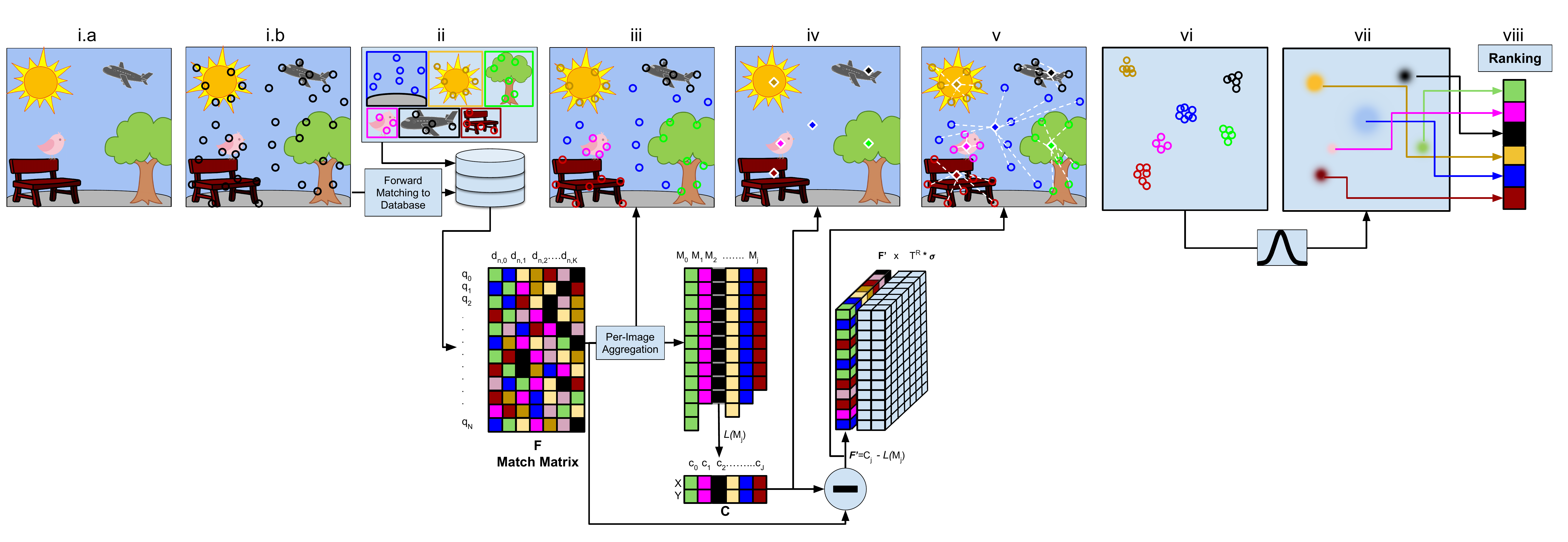}

   \caption{A more detailed depiction of the different steps of the OS2OS scoring algorithm.  (i.a) Query image $Q$. (i.b) $N$ Local features with associated geometric data (\textit{i.e.}, coordinates, scale, and rotation) are extracted from the query. (ii) A database containing images $[P_0,P_1,P_2,...,P_J]$ is collected. These images may contain donor objects to $Q$. (iii) Query features are assigned to $K$ nearest database matches in an $N \times K$ match matrix $F$. (iv) Matches to $J$ unique images are aggregated in sets $M_j, j \in J$. This is done using a parallel bin count. A feature center $c_j$ is estimated for each set of keypoints from a unique database image. (v) Keypoint transformations are calculated relative to the estimated centers. (vi) Keypoint transforms are applied to the database meta-features, projecting them into a voting space. (vii) A Probability Density Function is applied to the voting space to calculate cluster fitness. (viii) As each cluster represents an object, image scores are calculated on an object-by-object basis. Steps (ii) through (vi) are vectorized by reshaping all meta-feature data and match scores to conform to the dimensions of the match matrix $F$.  While slightly more memory intensive, each operation can be performed as either an addition or multiplication on coordinates of the match matrix.}
\label{fig:pipeline}
\end{figure*}
 
\section{Ablation of Parameters}

To verify the efficacy of different OS2OS algorithm components, we provide an ablation study on the different parts of the proposed method.  We first verify that centroid utilization does in fact contribute to higher retrieval performance.  We then interatively add in the Centrality Score (Eq.~\ref{eq:cs}), then the Angle Score (Eq.~\ref{eq:as}), then the logrithmic normalization (shown in Eq.~\ref{eq:score}).  These experiments are performed on the Nimble2018 dataset, as to reflect the algorithm's ability to accurately retrieve donor images when contributing only relatively small amounts of content to a composite query. The study results in Table~\ref{tab:ablation} show that each component provides at least a marginal improvement in retrieval performance, measured as Mean Average Precision (MAP). The results also show the relative stability of results despite changing Hough bin window sizes (WS, Eq.~\ref{eq:ws}).  This suggests that as long as window sizes divides the vote space by at least 5, we are able to effectively recover local objects and score them appropriately.
\begin{table*}[]
\begin{tabular}{|l|lllllllllllllll}
\hline
\multicolumn{1}{|l|}{}              & \multicolumn{3}{l|}{Pure Hough}                                                        & \multicolumn{3}{l|}{+ Centroid (Eq. \ref{eq:centroid})}                                               & \multicolumn{3}{l|}{CS (Eq. \ref{eq:cs})}                                                           & \multicolumn{3}{l|}{+ AS (Eq. \ref{eq:as})}                                                         & \multicolumn{3}{l|}{+ $log|O|$ (Eq. \ref{eq:score})}                                                              \\ \hline
\rowcolor[HTML]{C0C0C0} 
$ws_b$ (Eq. \ref{eq:bin})                                 & 5                         & 10                        & 20                        & 5                         & 10                        & 20                        & 5                         & 10                        & 20                        & 5                         & 10                        & 20                        & 5                         & 10                        & 20                                 \\ \hline
\multicolumn{1}{|l|}{R@k=200} & \multicolumn{1}{l|}{76.2} & \multicolumn{1}{l|}{76.3} & \multicolumn{1}{l|}{76.2} & \multicolumn{1}{l|}{77.5} & \multicolumn{1}{l|}{78.1} & \multicolumn{1}{l|}{77.5} & \multicolumn{1}{l|}{80.2} & \multicolumn{1}{l|}{80.9} & \multicolumn{1}{l|}{79.3} & \multicolumn{1}{l|}{81.2} & \multicolumn{1}{l|}{82.1} & \multicolumn{1}{l|}{82.0} & \multicolumn{1}{l|}{83.1} & \multicolumn{1}{l|}{83.2} & \multicolumn{1}{l|}{\textbf{83.4}} \\ \hline
\end{tabular}
\caption{
Ablation study iteratively adding in algorithm components, with voting window sizes (WS) of 5, 10, and 20 for each experiment. Columns 1-3 (Pure Hough) perform simple summed Hough voting (without further scoring) within the given Hough bins.
}
\label{tab:ablation}
\vspace{-0.2cm}
\end{table*}

\subsection{Further study of weighted centroid use}
\begin{figure}[]
\hspace*{-1.3cm}  
\centering
   \includegraphics[width=.8\linewidth]{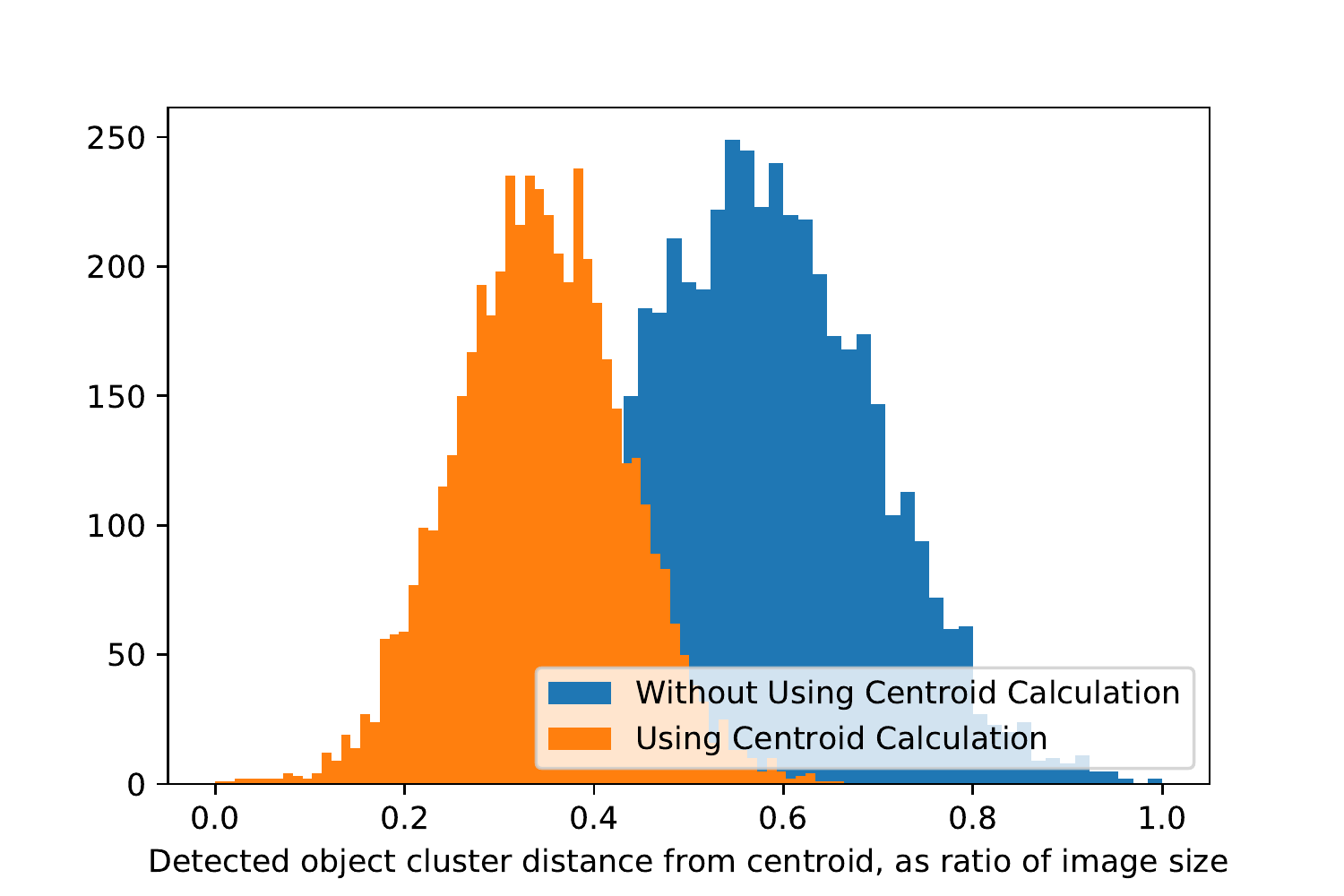}

   \caption{Comparison of Distribution of L2 distances of Hough vote vectors. When not using centroid calculation (blue), we simply set the vote point to be the center of the query image's dimensions. Results show that vote distances are significantly shorter when using weighted centroid voting, which helps to reduce the magnification of noisy voting.}
\label{fig:centroids}
\end{figure}

This section aims to show the benefit of using the weighted centroid calculated from matched keypoint locations as the voting center for given images.  We perform two matching experiments using a random sample of 5000 images within the Google Landmarks dataset, and extract the average vector length of each vote projection (Eq.~\ref{eq:totalT}).  Our analysis in Figure~\ref{fig:centroids} shows that utilizing this weighted average.
\\
\\
\section{Qualitative Results}
Here we provide more visual results of the NH score discovering objects in datasets. We provide visual representations of query cases from all 3 datasets used within the paper, showing success and failure cases for the Nimble, Google, and Reddit datasets.  In each figure, the first row corresponds to the highest voted center of points of interest in the query by the points detected in the corresponding rank retrieved image. The second row shows the heatmap of the probability distribution of votes of matched features to the query for the particular image. The third row contains the original rank retrieved image. The columns correspond to the different ranks. Because we use weighted centroid voting, we can visualize where the algorithm matches objects between the query and retrieved images by overlaying the generated Probability Density Function (PDF) of Hough votes on top of each retrieved image.
\vspace{-0.5cm}
\begin{figure*}[]
\centering
   \includegraphics[width=0.85\linewidth]{Figures/f9e538ea76c5b551.pdf}
   \includegraphics[width=0.85\linewidth]{Figures/f149998853b42e45.pdf}
   \includegraphics[width=0.85\linewidth]{Figures/87e393dffdf4f517.pdf}
   \caption{Results of 3 queries from the Google Landmarks dataset.}
\label{fig:googlelandmarks}
\end{figure*}

\begin{figure*}[]
\centering
   \includegraphics[width=0.9\linewidth]{Figures/e6cdf2e84ea356ca.pdf}
   \caption{A failure scenario from the Google Landmarks dataset where the matches do not appear to be coherent and high affinity matches are false positives.}
\label{fig:googlelandmarksfailure}
\end{figure*}

\begin{figure*}[]
\centering
   \includegraphics[width=0.85\linewidth]{Figures/g1245.pdf}
   \caption{Example of a query result from the Reddit Provenance Dataset. Incorrect match examples are highlighted in red text. Note the Rank 6 and Rank 8 matches, in which multiple objects from the query are independently donated to the matched result.  The heat maps reflect the multiple object matching.}
   \includegraphics[width=0.85\linewidth]{Figures/g1079.pdf}
   \caption{Example of a query result from the Reddit Provenance Dataset. Incorrect match examples are highlighted in red text. Of particular interest is the Rank 5 match, in which the query donates the same object multiple times to the result. The discovery of these 4 donations is reflected in the heat map.}
\label{fig:googlelandmark3}
\end{figure*}

\begin{figure*}[]
\centering
   \includegraphics[width=0.85\linewidth]{Figures/2fbde6ce8685469be654ace559df2483.pdf}
   \caption{Example of a query result from the Nimble Provenance Dataset. Incorrect match examples are highlighted in red text. In this case, the search algorithm finds both the Man and the puddle as donated objects within the query.}
   \includegraphics[width=0.85\linewidth]{Figures/26ac633432d9f4a689f5c6a05d9bfa26.pdf}
   \caption{Example of a query result from the Nimble Provenance Dataset. Incorrect match examples are highlighted in red text. The Rank 3 match actually appears within the window frame of the query image.}
\label{fig:googlelandmark3}
\end{figure*}